\documentclass[acmlarge,11pt, nonacm]{acmart}
\usepackage{booktabs}    
\usepackage{colortbl}    
\usepackage{array}       
\usepackage{multirow}
\usepackage{xspace}
\usepackage{listings}
\usepackage{subcaption}
\usepackage{tikz}
\usepackage{paralist}
\usepackage{cleveref}
\usepackage{tcolorbox}
\usepackage{xcolor}
\usepackage{amsthm}
\usepackage{thmtools}

\newcommand{\llmic}{LLMic\xspace}

\definecolor{gray}{HTML}{ededed}
\colorlet{shadecolor}{gray}
\tcbset{
	colback=gray,
	boxrule=0.5pt,
	boxsep=1pt,
	left=3pt,
	right=3pt,
	top=3pt,
	bottom=3pt,
	sharp corners,
	colframe=black,
}

\declaretheoremstyle[
    spaceabove=6pt, 
    spacebelow=6pt,
    headfont=\normalfont\bfseries,
    notefont=\mdseries, 
    notebraces={(}{)},
    bodyfont=\normalfont,
    postheadspace=1em,
    headpunct={:},
    headformat=\NAME\NOTE,
]{defstyle}

\title{\llmic: Romanian Foundation Language Model}

\author{Vlad-Andrei Bădoiu}
\affiliation{
    \institution{University Politehnica of Bucharest}
    \city{Bucharest}
    \country{Romania}
}

\authornote{These authors contributed equally to this work.}

\author{Alexandru M. Gherghescu}
\affiliation{
    \institution{University Politehnica of Bucharest}
    \city{Bucharest}
    \country{Romania}
}
\authornotemark[1]

\author{Alexandru Agache}
\affiliation{
    \institution{University Politehnica of Bucharest}
    \city{Bucharest}
    \country{Romania}
}

\author{Mihai-Valentin Dumitru}
\affiliation{
    \institution{University Politehnica of Bucharest}
    \city{Bucharest}
    \country{Romania}
}

\author{Costin Raiciu}
\affiliation{
  \institution{University Politehnica of Bucharest}
   \institution{Broadcom Inc.}
    \city{Bucharest}
    \country{Romania}
}

\makeatletter
\let\@authorsaddresses\@empty
\makeatother

\begin{abstract}
Recent advances in Large Language Models (LLMs) have demonstrated remarkable
capabilities across various tasks with commercial models leading the way. While
open models usually operate at a smaller scale, they maintain competitiveness
through specialization and fine-tuning. However, a significant challenge
persists: open models often underperform in low-resource languages due to
limited representation in the training corpus.

In this paper, we present \llmic, a
bilingual foundation language model designed specifically for the Romanian
Language. We document the complete process of pretraining a foundation model
for a low-resource language, including corpus construction, architecture
selection, and hyper-parameter optimization. Our evaluation demonstrates that
\llmic can be specialized for tasks in the target
language, achieving results comparable to other much larger open models. We
show that fine-tuning \llmic for language translation after the initial
pretraining phase outperforms existing solutions in English-to-Romanian
translation tasks. This opens the path for efficient large-scale processing
for the Romanian language community, using the much smaller \llmic model.

\end{abstract}

\begin{document}

\maketitle

\section{Introduction}

The emergence of Large Language Models (LLMs) has transformed the landscape of
natural language processing. We are witnessing an ever-growing number of open
and proprietary models.
Leveraging hyperscale computing infrastructure, commercial models such as
GPT-4, Claude, and Gemini attained the trillion-parameters
threshold~\cite{decoder_gpt4_2024} achieving never seen before
capabilities~\cite{brown2020language, wei2022emergent}. At this size, models
are great general purpose problem-solvers, demonstrating sophisticated
reasoning.

On the open models front, despite operating at significantly lower parameter
counts, we have a vibrant ecosystem~\cite{minaee2024large}. Models such as
Llama ~\cite{dubey2024llama}, Deepseek~\cite{deepseek}, Qwen~\cite{bai2023qwen}
and Mistral~\cite{jiang2023mistral} are able to be competivitve on a variety of
tasks~\cite{chiang2024chatbot}. Through fine-tuning techniques, these open
models can be adapted further, for state-of-the-art task-specific performance. 
The advantages of open models are multiple: reduced operational costs, efficient
processing of large-scale data through specialized compact models,
privacy through self-deployment, and low memory footprint enabling running on
edge devices such as smartphones.

A significant challenge in language model development lies in achieving
consistent performance across diverse languages. Open-source models,
constrained by their reduced parameter counts and English-centric pretraining
datasets, typically underperform on low-resource languages compared to
commercial alternatives~\cite{ali2024survey}. While open models leverage public datasets like Dolma~\cite{dolma} and
FineWebEdu~\cite{fineweb}, such collections predominantly target widely spoke
languages , with limited representation for other languages.
While fine-tuning can enhance a model's capabilities in specific
languages~\cite{masala2024vorbecsti, garcia2024introducing}, the results are
underwhelming compared to the performance in languages that dominate the
pretraining data. While fine-tuning can be used to improve the capabilities in
specific languages~\cite{masala2024vorbecsti, garcia2024introducing}
performance, it does provide underwhelming results compared to the languages
that make most of the pretraining dataset.

However, recent research has shown that it is possible to classify an LLM's
neurons into ``language-specific'' and ``language-agnostic''
categories~\cite{tang2024language}, whith language specific-specific neurons
being only a tiny fraction of the entire model. The authors of
\cite{tang2024language} go as far as showing that, through manual intervention
in the LLM's inference process, activations of language-specific neurons can be
modified to steer the model towards a particular language. This finding has
important implications: it suggests that strong cross-lingual performance on
non-linguistic tasks might be achievable by combining a comprehensive English
``knowledge corpus'' with relatively small amounts of language-specific
documentation.



Our work focuses on Romanian, a language that
represents approximately 0.6\% of Common Crawl
pages~\cite{commoncrawl_languages} and where existing open models demonstrate
limited capabilities~\cite{masala2024vorbecsti}. To address this gap, we
introduce \llmic, a 3B-parameter bilingual Romanian-English foundation model.
This paper presents a comprehensive study on pretraining a foundation model; we
address critical challenges, beginning with the construction of high-quality
pretraining corpora for the Romanian language - a resource notably absent from the
state-of-the-art open datasets~\cite{fineweb, weber2024redpajama, dolma,
penedo2023refinedweb, li2024datacomp}. Next, we discuss the process of selecting
the model's
architecture, optimization of hyperparameters, and development of an efficient
runtime configuration. To showcase the usefulnes of \llmic, we fine-tune the
model for translation tasks and evaluate its performance against existing
models, including fine-tuned model for the Romanian language. 

By releasing
\llmic to the research community, we aim to bootstrap the development of
natural language processing tools and models for the Romanian language.
LLMic\footnote{https://huggingface.co/faur-ai/LLMic} is released under the Apache 2.0 License.

\section{Pretraining Data}

Data scarcity presents a fundamental challenge in developing models for
Romanian language processing. Acquiring hundreds of billions of tokens requires
extensive filtering and cleaning of CommonCrawl's petabyte-scale dataset.
Obtaining additional curated content often necessitates applying optical
character recognition (OCR)~\cite{TessOverview} to public domain documents, or
parsing complex formats (HTML, EPUB) to extract text.

We trained \llmic on a mixture of filtered web data and curated sources in both
Romanian and English. The data is split in 300B tokens for Romanian and 700B
tokens for English, as shown in Table~\ref{tab:dataset_composition}.
For the English subset of the pretraining data, we
use FineWebEdu~\cite{fineweb} due to its LLM-assisted quality filtering, and
we further source a curated set of documents from Dolma~\cite{dolma}; for
Romanian, we use the entirety of FuLG~\cite{buadoiu2024fulg}, and we further repeat a small number of
high-quality documents. We mask all personal identifiable information
(PII). A detailed overview of the dataset is discussed below.

\begin{table}[htbp]
    \centering
    \begin{tabular}{lr}
        \toprule
        \textbf{Source} & \textbf{Size}\\
        \midrule
        \multicolumn{2}{l}{\textit{Romanian (300B)}} \\
        \quad Web Sources & 621 GB \\
        \quad Discussions, Curated \& Parallel & 10 GB \\
        \midrule
        \multicolumn{2}{l}{\textit{English (700B)}} \\
        \quad FineWebEdu & -- \\
        \quad Dolma Subset & 109 GB \\
        \midrule
        \bottomrule
    \end{tabular}
    \caption{Dataset Composition by Language and Source}
    \label{tab:dataset_composition}
\end{table}

\textbf{Web Sources.} We leverage two filtered CommonCrawl sources for Romanian
language data: FuLG~\cite{buadoiu2024fulg}, extracted using the CCNet
pipeline~\cite{ccnet}, and contains 220B tokens when processed with the Llama
tokenizer; and mC4~\cite{raffel2020exploring}, a multilingual cleaned version
of CommonCrawl, comprising 42B Romanian tokens. While they share the source,
the two corpora differ in their language detection algorithms and cleaning and
deduplication techniques. To remove duplicates resulted from the merge, we
apply both fuzzy and exact deduplication using the
RedPajama~\cite{weber2024redpajama, redpajama} pipeline. We further augment our dataset
by incorporating filtered content from recent CommonCrawl dumps (2024-22,
2024-26, 2024-30, and 2024-33).

\paragraph{\textbf{Curated Data.}}
We extend our Romanian web corpus with curated data from multiple sources,
including the Romanian Wikipedia and public documents from Romanian
institutions processed. For the English portion, we utilize a curates subset
Dolma~\cite{dolma}, specifically incorporating content from books, Wikipedia,
and research papers.

\paragraph{\textbf{Discussions.}} Our dataset includes a subset of discussions
from public forums. For Romanian content, we utilized forum data from
the the Pushshift~\cite{baumgartner2020pushshift} dataset. The English forum discussions
were sourced from the Dolma dataset.

\paragraph{\textbf{Parallel Data.}}
We incorporated parallel Romanian-English data from multiple sources, including
translated official documents from the European Union and the
ParaCrawl~\cite{banon2020paracrawl} parallel corpus.

\section{Tokenizer}

\begin{figure}[htbp]
    \centering
    \includegraphics[width=0.65\textwidth]{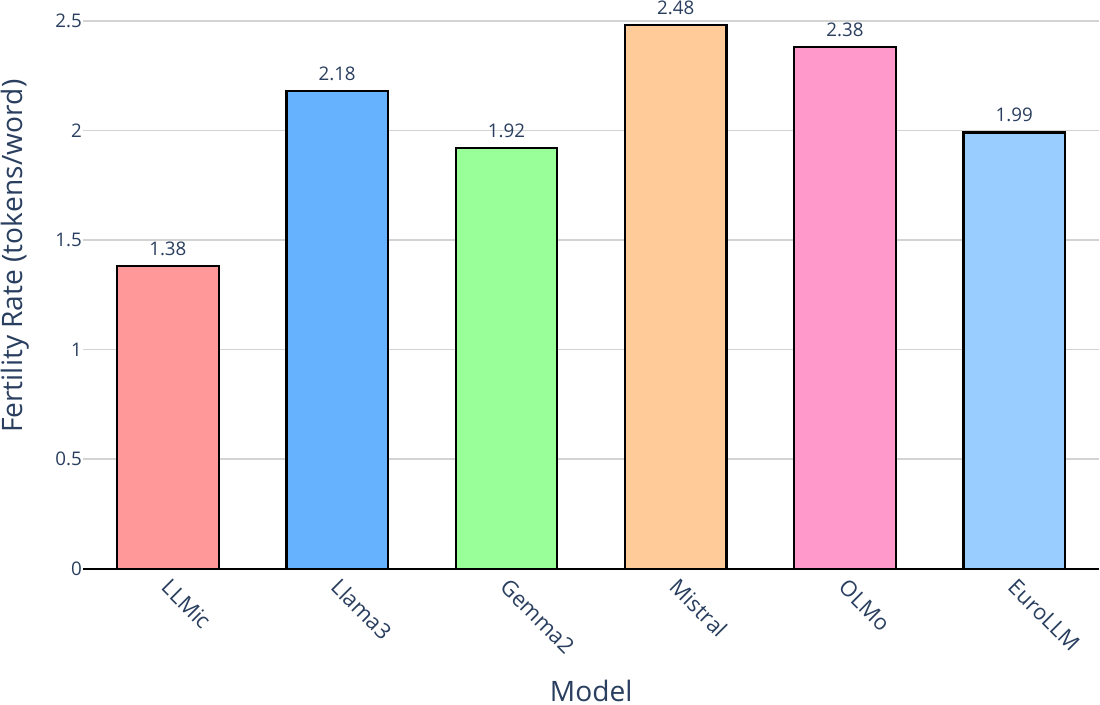}
    \caption{Tokenizer fertility analysis across different languages and text
    sources. The graph shows the relationship between input text and resulting
    tokens.}
    \label{fig:tokenizer_fertility}
\end{figure}

A model's tokenizer can play a role in both model inference speed and the overall
capabilities of the model.
As shown in Figure~\ref{fig:tokenizer_fertility}, existing
popular tokenizers exhibit poor compression rates for Romanian text.

To address this limitation, we built a BPE tokenizer based on the GPT-NeoX
architecture~\cite{black2022gpt}, training it on 7B tokens equally distributed
between FineWebEdu~\cite{fineweb} and our Romanian dataset. Our implementation
uses an uncased vocabulary of 128,000 tokens and intentionally omits the
Romanian characters: ă, â, î, ș, ț, replacing them with  a, i, s, t respectively
-- a decision based on preliminary experiments where these characters degraded
model performance. This optimized design achieves a fertility rate of
1.38, substantially improving upon existing tokenizers.

\section{Architecture}

The architecture of \llmic is a classical decoder-only
transformer~\cite{vaswani2017attention}. Following an approach similar to
OLMo~\cite{groeneveld2024olmo}, we surveyed the leading families of models at
this time, including Gemma~\cite{team2024gemma},
OLMo~\cite{groeneveld2024olmo}) and Llama2~\cite{touvron2023llama}). Based on
this, we selected Llama2 as the underlying architecture. All the details about
\llmic can be consulted in Table~\ref{tab:model-arch}. We use grouped query
attention (GQA)~\cite{ainslie2023gqa}, for faster inference speed, rotary
positional embeddings (RoPE)~\cite{su2024roformer} to enable context extension
and SiLU activation~\cite{elfwing2018sigmoid}. 

\begin{table}[htbp]
\centering
\begin{subtable}[t]{0.48\textwidth}
    \centering
    \begin{tabular}{lr}
    \toprule
    \textbf{Parameter} & \textbf{Value} \\
    \midrule
    Sequence Length & 2048 \\
    Number of Layers & 24 \\
    Embedding Size & 2,560 \\
    FFN Hidden Size & 10,240 \\
    Number of Heads & 20 \\
    Number of KV Heads & 5 \\
    Activation Function & SiLU \\
    Position Encodings & RoPE ($\Theta=500{,}000$) \\
    Layer Norm & RMSNorm ($\epsilon=10^{-5}$) \\
    Tied Embeddings & No \\
    \bottomrule
    \end{tabular}
    \caption{Model Architecture}
    \label{tab:model-arch}
\end{subtable}
\hfill
\begin{subtable}[t]{0.48\textwidth}
    \centering
    \begin{tabular}{lr}
    \toprule
    \textbf{Parameter} & \textbf{Value} \\
    \midrule
    Batch Size (per GPU) & 8 \\
    Warmup & 3000 steps \\
    Gradient Accumulation & 2 \\
    Sequence Length & 2048 \\
    Weight Decay & 0.1 \\
    Learning Rate Scheduler & Cosine with Min LR \\
    Learning Rate & 4 × 10\textsuperscript{-4} \\
    Adam $\beta$\textsubscript{1} & 0.9 \\
    Adam $\beta$\textsubscript{2} & 0.95 \\
    Adam $\epsilon$ & 1 × 10\textsuperscript{-5} \\
    \bottomrule
    \end{tabular}
    \caption{Training Hyper-parameters}
    \label{tab:hyperparams}
\end{subtable}
\caption{Model Specifications and Training Parameters}
\label{tab:combined}
\end{table}

\section{Training}

We analyzed existing model's training regimes, as well as conducted a few
learning rate ablations on our own, in order to find the (close to) optimal
learning rate. Generally speaking, we followed the rule-of-thumb:
increase the learning rate as much as you can, as long as the training loss
doesn't diverge.

Our initial experiments with a constant learning rate of $4 \times 10^{-3}$
proved to be excessive, as evidenced by the model's outputs becoming overly
repetitive and showing signs of text memorization. Subsequently, we implemented
a cosine learning rate scheduler with a maximum learning rate of $4 \times
10^{-3}$. The complete set of hyper-parameters is detailed in
Table~\ref{tab:hyperparams}.

For the distributed training, we implemented a custom framework leveraging
Hugging Face's Transformers~\cite{wolf2020transformers}. The training process
utilized Fully Sharded Data Parallel (FSDP)~\cite{zhao2023pytorch} parallelism
with full sharding. To optimize performance, we used \texttt{bfloat16} mixed
precision training alongside the optimized Liger kernels~\cite{hsu2024liger}.

The training procedure followed a multi-phase approach. During the initial 50B
tokens, we used a 50:50 split between Romanian and English, as we reasoned the
model should first learn both the languages equally. Subsequently, we used the
30:70 Romanian-English split, as dictated by the dataset. Towards the end of
the pretraining phase, we re-used high-quality Romanian documents multiple
times.

\section{Evaluation}



We developed \llmic as a foundation model optimized for Romanian language
processing tasks. To assess its capabilities, we evaluated its performance on
English-to-Romanian translation by fine-tuning the model and testing it on the
WMT~\cite{bojar2016findings} translation benchmark. Since \llmic operates on
uncased text without diacritics, we modified the WMT data by removing
diacritics and converting to lowercase, while leaving everything else
unchanged.


\begin{table}[htbp]
\centering
\begin{subtable}[t]{0.45\textwidth}
\centering
\renewcommand{\arraystretch}{1.2}
\begin{tabular}{l r}
\toprule
\textbf{Method} & \textbf{Score} \\
\midrule
Full Precision (FP16) & 41.01 \\
GPTQ INT8 (calibrated on C4) & 41.01 \\
GPTQ INT4 (calibrated on C4) & 40.29 \\
BitsAndBytes INT4 & 40.44 \\
LoRA (r = 8) & 37.05 \\
\bottomrule
\end{tabular}
\caption{Quantization Methods Performance}
\label{tab:quant-methods}
\end{subtable}
\hfill
\begin{subtable}[t]{0.45\textwidth}
\centering
\renewcommand{\arraystretch}{1.2}
\begin{tabular}{l r}
\toprule
\textbf{Model} & \textbf{Score} \\
\midrule
\llmic & 41.01 \\
  mBART~\cite{liu2020multilingual} & 38.50 \\
Llama-3.1-8B-Instruct & 29.02 \\
RoMistral-7b-Instruct & 27.70 \\
RoLlama3-8b-Instruct & 27.31 \\
Mistral-7B-Instruct-v0.2 & 26.19 \\
RoGemma-7b-Instruct & 25.96 \\
Gemma-1.1-7b-it & 25.48 \\
\bottomrule
\end{tabular}
\caption{Model Performance}
\label{tab:model-perf}
\end{subtable}
\caption{Performance Comparison of Quantization Methods and Language Models}
\label{tab:combined-comparison}
\end{table}

The results, presented in Table~\ref{tab:model-perf}, show that
\llmic outperforms both established open models like
LLaMA-2~\cite{touvron2023llama}, Mistral~\cite{jiang2023mistral}, and
Gemma~\cite{gemma}, as well as previous work on fine-tuning open models for
Romanian~\cite{masala2024vorbecsti}. Our qualitative analysis indicates that
\llmic achieves comparable translation quality to closed models such as
ChatGPT.

We designed \llmic with both at scale and on device processing in mind. To achieve this,
we looked into reducing its footprint even more by leveraging quantization.
This reduces the memory footprint and enables the use of lower lower accuracy,
but at much higher number of  floating point operations per second. Although
quantization results are highly dependant on the task at hand, we noticed only
a marginal impact on performance. Results are shown in
table~\ref{tab:quant-methods}.

Through the above results, we showcase \llmic as an effective solution for edge
computing and large-scale processing. We hope such approaches of using smaller
models for specific tasks can inspire the community to build smaller, more
specialized models, which in turn can be used by many more individuals without
the need for expensive resources.



\bibliographystyle{ACM-Reference-Format}
\bibliography{ref}

\end{document}